\documentclass[letterpaper, 10 pt, conference]{ieeeconf}  % Comment this line out if you need a4paper

\IEEEoverridecommandlockouts                              % This command is only needed if
\usepackage{amsmath,amssymb,amsfonts}
\usepackage{algpseudocode,algorithm}
\usepackage{subcaption}
\usepackage{graphicx}
\usepackage{cite}

\title{\LARGE \bf
Consolidated Adaptive T-soft Update for Deep Reinforcement Learning
}

\author{Taisuke Kobayashi$^{1}$% <-this % stops a space
\thanks{$^{1}$Taisuke Kobayashi is with the Division of Information Science, Nara Institute of Science and Technology, 8916-5 Takayama-cho, Ikoma, Nara 630-0192, Japan
        {\tt\small kobayashi@is.naist.jp}}%
}

\begin{document}

\maketitle
\thispagestyle{empty}
\pagestyle{empty}

%%%%%%%%%%%%%%%%%%%%%%%%%%%%%%%%%%%%%%%%%%%%%%%%%%%%%%%%%%%%%%%%%%%%%%%%%%%%%%%%
% no more than 200 words
\begin{abstract}

Demand for deep reinforcement learning (DRL) is gradually increased to enable robots to perform complex tasks, while DRL is known to be unstable.
As a technique to stabilize its learning, a target network that slowly and asymptotically matches a main network is widely employed to generate stable pseudo-supervised signals.
Recently, T-soft update has been proposed as a noise-robust update rule for the target network and has contributed to improving the DRL performance.
However, the noise robustness of T-soft update is specified by a hyperparameter, which should be tuned for each task, and is deteriorated by a simplified implementation.
This study develops adaptive T-soft (AT-soft) update by utilizing the update rule in AdaTerm, which has been developed recently.
In addition, the concern that the target network does not asymptotically match the main network is mitigated by a new consolidation for bringing the main network back to the target network.
This so-called consolidated AT-soft (CAT-soft) update is verified through numerical simulations.

\end{abstract}

%%%%%%%%%%%%%%%%%%%%%%%%%%%%%%%%%%%%%%%%%%%%%%%%%%%%%%%%%%%%%%%%%%%%%%%%%%%%%%%%

%%%%%%%%%%%%%%%%%%%%%%%%%%%%%%%%%%%%%%%%%%%%%%%%%%%%%%%%%%%%%%%%%%%%%%%%%%%%%%%%
\section{Introduction}

%%% robotics for complicated tasks
As robotic and machine learning technologies remarkably develop, the tasks required of intelligent robots becomes more complex: e.g.
physical human-robot interaction~\cite{modares2015optimized,kobayashi2021whole};
work on disaster sites~\cite{kobayashi2015selection,delmerico2019current};
and manipulation of various objects~\cite{tsurumine2019deep,kroemer2021review}.
In most cases, these complex tasks have no accurate analytical model.
To resolve this difficulty, deep reinforcement learning (DRL)~\cite{sutton2018reinforcement} has received a lot of attention as an alternative to classic model-based control.
Using deep neural networks (DNNs) as a nonlinear function approximator, DRL can learn the complicated policy and/or value function in a model-free manner~\cite{mnih2015human,haarnoja2018soft}, or learn the complicated world model for planning the optimal policy in a model-based manner~\cite{chua2018deep,okada2020planet}.

%%% target network
Since DNNs are nonlinear and especially model-free DRLs must generate pseudo-supervised signals by themselves, making DRL unstable.
Techniques to stabilize learning have been actively proposed, such as the design of regularization~\cite{tsurumine2019deep,haarnoja2018soft,kobayashi2021proximal} and the introduction of the model that makes learning conservative~\cite{osband2016deep,kobayashi2019student}.
Among them, a target network is one of the current standard techniques in DRL~\cite{mnih2015human,kobayashi2021t}.
After generating it as a copy of the main network to be learned by DRL, an update rule is given to make it slowly match the main network at regular intervals or asymptotically.
In this case, the pseudo-supervised signals generated from the target network are more stable than those generated from the main network, which greatly contributes to the overall stability of DRL.

%%% t-soft update
The challenge in using the target network is its update rule.
It has been reported that too slow update stagnates the whole learning process~\cite{kim2019deepmellow}, while too fast update reverts to instability of the pseudo-supervised signals.
A new update rule, T-soft update~\cite{kobayashi2021t}, has been proposed to mitigate the latter problem.
This method provides a mechanism to limit the amount of updates when the main network deviates unnaturally from the target network, which can be regarded as noise.
Such noise robustness enables to stabilize the whole learning process even with the high update rate by appropriately ignoring the unstable behaviors of the main network.

%%% problem
However, the noise robustness of T-soft update is specified as a hyperparameter, which must be set to an appropriate value depending on the task to be solved.
In addition, the simplified implementation for detecting noise deteriorates the noise robustness.
More sophisticated implementation with the adaptive noise robustness is desired.
As another concern, when the update of the target network is restricted like T-soft update does, the target network may not asymptotically match the main network.
A new constraint is needed to avoid the situation where the main network deviates from the target network.

%%% proposal
Hence, this paper proposes two methods to resolve each of the above two issues: i) the adaptive and sophisticated implementation of T-soft update and; ii) the appropriate consolidation of the main target network to the target network.
Specifically, for the issue i), a new update rule, so-called adaptive T-soft (AT-soft) update, is developed based on the recently proposed AdaTerm~\cite{ilboudo2022adaterm} formulation, which is an adaptively noise-robust stochastic gradient descent method.
This allows us to sophisticate the simplified implementation of T-soft update and improve the noise robustness, which can be adaptive to the input patterns.
For the issue ii), a new consolidation so that the main network is regularized to the target network when AT-soft update restricts the updates of the target network.
By implementing it with interpolation, the parameters in the main network that naturally deviate significantly from that of the target network, are updated to a larger extent.
With this consolidation, the proposed method is so-called consolidated AT-soft (CAT-soft) update.

%%% results
To verify CAT-soft update, typical benchmarks implemented by Pybullet~\cite{coumans2016pybullet} are tried using the latest DRL algorithms~\cite{kobayashi2021proximal,schaul2015prioritized}.
It is shown that even though the learning rate is larger than the standard value for DRL, the task performance is improved by CAT-soft update and more stable learning can be achieved.
In addition, the developed consolidation successfully suppresses the divergence between the main and target networks.

%%%%%%%%%%%%%%%%%%%%%%%%%%%%%%%%%%%%%%%%%%%%%%%%%%%%%%%%%%%%%%%%%%%%%%%%%%%%%%%%
\section{Preliminaries}

%%%%%%%%%%%%%%%%%%%%%%%%%%%%%%%%%%%%%%%%
\subsection{Reinforcement learning}

First of all, the basic problem statement of DRL and an actor-critic algorithm, which can handle continuous action space, is natural choice~\cite{sutton2018reinforcement} as one of the basic algorithms for robot control.
Note that the proposed method can be applied to other algorithms with the target network.

In DRL, an agent interacts with an unknown environment under Markov decision process (MDP) with the current state $s$, the agent's action $a$, the next state $s^\prime$, and the reward from the environment $r$.
Specifically, the environment implicitly has its initial randomness $p_0(s)$ and its state transition probability $p_e(s^\prime \mid s, a)$.
Since the agent can act on the environment's state transition through $a$, the goal is to find the optimal policy to reach the desired state.
To this end, $a$ is sampled from a state-dependent trainable policy, $\pi(a \mid s; \theta_{\pi})$, with its parameters set $\theta_{\pi}$ (a.k.a. weights and biases of DNNs in DRL).
The outcome of the interaction between the agent and the environment can be evaluated as $r = r(s, a, s^\prime)$.

By repeating the above process, the agent gains the sum of $r$ over the future (so-called return), $R = \sum_{k=0}^\infty \gamma^k r_k$, with $\gamma \in [0, 1)$ discount factor.
The main purpose of DRL is to maximize $R$ by optimizing $\pi$ (i.e. $\theta_{\pi}$).
However, $R$ cannot be gained due to its future information, hence its expected value is inferred as a trainable (state) value function, $V(s; \theta_{V}) = \mathbb{E}[R \mid s]$, with its parameters set $\theta_{V}$.
Finally, DRL optimizes $\pi$ to maximize $V$ while increasing the accuracy of $V$.

To learn $V$, a temporal difference (TD) error method is widely used as follows:
\begin{align}
    \mathcal{L}_{V}(\theta_{V}) &= \cfrac{1}{2} (y - V(s; \theta_{V}))^2
    \label{eq:loss_value} \\
    y &= r + \gamma V(s^\prime; \bar{\theta}_{V})
\end{align}
where $y$ denotes the pseudo-supervised signal generated from the target network with the parameters set $\bar{\theta}_{V}$ (see later).
By minimizing $\mathcal{L}_{V}$, $\theta_{V}$ can be optimized to correctly infer the value over $s$.

To learn $\pi$, a policy-gradient method is applied as follows:
\begin{align}
    \mathcal{L}_{\pi}(\theta_{\pi}) &= - (y - V(s; \theta_{V})) \cfrac{\pi(a \mid s; \theta_{\pi})}{b(a \mid s; \bar{\theta}_{\pi})}
    \label{eq:loss_policy}
\end{align}
where $a$ is sampled from the alternative policy $b$, which is often given by the target network with the parameters set $\bar{\theta}_{\pi}$.
The sampler change is allowed by the importance sampling, and the likelihood ratio is introduced as in the above loss function.
By minimizing $\mathcal{L}_{\pi}$, $\theta_{\pi}$ can be optimized to reach the state with higher value.

%%%%%%%%%%%%%%%%%%%%%%%%%%%%%%%%%%%%%%%%
\subsection{Target network with T-soft update}

The target network with $\bar{\theta}_{V, \pi}$ is briefly introduced together with the latest update rule, T-soft update~\cite{kobayashi2021t}.
First, with the initialization phase of the main network, the target network is also created as a copy with $\bar{\theta}_{V, \pi} = \theta_{V, \pi}$.
Since the copied $\bar{\theta}_{V, \pi}$ is given independently of $\theta_{V, \pi}$, and not updated through the minimization problem of eqs.~\eqref{eq:loss_value} and~\eqref{eq:loss_policy}.
Therefore, the pseudo-supervised signal $y$ has the same value for the same input, which greatly contributes to the stability of learning by making the minimization problem stationary.

However, in practice, if $\bar{\theta}_{V, \pi}$ is fixed at its initial value, the correct $y$ cannot be generated and the task is never accomplished.
Thus, $\bar{\theta}_{V, \pi}$ must be updated slowly towards $\theta_{V, \pi}$ as in alternating optimization~\cite{bezdek2003convergence}.
When the target network was first introduced, a technique called \textit{hard update} was employed~\cite{mnih2015human}, where $\theta_{V, \pi}$ was updated a certain number of times and then copied again as $\bar{\theta}_{V, \pi} = \theta_{V, \pi}$.
Afterwards, \textit{soft update} shown in the following equation has been proposed to make $\bar{\theta}_{V, \pi}$ asymptotically match $\theta_{V, \pi}$ more smoothly.
\begin{align}
    \bar{\theta}_{V, \pi} \gets (1 - \tau) \bar{\theta}_{V, \pi} + \tau \theta_{V, \pi}
    \label{eq:tar_soft}
\end{align}
where $\tau \in (0, 1]$ denotes the update rate.

The above update rule is given as an exponential moving average, and all new inputs are treated equivalently.
As a result, even when $\theta_{V, \pi}$ is incorrectly updated, its adverse effect as noise is reflected into $\bar{\theta}_{V, \pi}$.
This effect is more pronounced when $\tau$ is large, but mitigating this by reducing $\tau$ causes a reduction in learning speed~\cite{kim2019deepmellow}.

To tackle this problem, T-soft update that is robust to noise even with relatively large $\tau$ has recently been proposed~\cite{kobayashi2021t}.
It regards the exponential moving average as update of a location parameter of normal distribution, and derives a new update rule by replacing it with student-t distribution that is more robust to noise by specifying degrees of freedom $\nu \in \mathbb{R}_{+}$.
T-soft update is described in Alg.~\ref{alg:tsoft}.
Note that $\sigma^2$ and $W$ must be updated as new internal states.

With mathematical explanations, the issues of T-soft update are again summarized as below.
\begin{enumerate}
    \item Since $\nu$ must be specified as a constant in advance, it must be tuned for each task to provide the appropriate noise robustness to maximize performance.
    \item The larger $\Delta_i \sigma_i^{-2}$ is, the more the update is suppressed.
    However, the simple calculation of $\Delta_i$ as mean square error makes it easier to hide the noise hidden in the $i$-th subset.
    \item If $\tau_i$ is frequently close to zero (i.e. no update), there is a risk that $\bar{\theta}_{V, \pi}$ will not asymptotically match $\theta_{V, \pi}$.
\end{enumerate}

%Algorithm
\begin{algorithm}[tb]
    \caption{T-soft update~\cite{kobayashi2021t}}
    \label{alg:tsoft}
    \begin{algorithmic}[1]
        \State{(Initialize $\bar{\theta}_i = \theta_i$, $W_i = (1 - \tau) \tau^{-1}$, $\sigma_i = \epsilon \ll 1$)}
        \For{$\theta_i, \bar{\theta}_i \subset \theta_{V, \pi}, \bar{\theta}_{V, \pi}$}
            \State{$\Delta_i^2 = \frac{1}{d_i} \sum_{j=1}^{d_i}(\theta_{i,j} - \bar{\theta}_{i,j})^2$}
            \State{$w_i = (\nu + 1)(\nu + \Delta_i^2 \sigma_i^{-2})^{-1}$}
            \State{$\tau_i = w_i (W_i + w_i)^{-1}$}
            \State{$\tau_{\sigma_i} = \tau w_i \nu (\nu + 1)^{-1}$}
            \State{$\bar{\theta}_i \gets (1 - \tau_i) \bar{\theta}_i + \tau_i \theta_i$}
            \State{$\sigma_i^2 \gets (1 - \tau_{\sigma_i}) \sigma_i^2 + \tau_{\sigma_i} \Delta_i^2$}
            \State{$W_i \gets (1 - \tau) (W_i + w_i)$}
        \EndFor
    \end{algorithmic}
\end{algorithm}

%%%%%%%%%%%%%%%%%%%%%%%%%%%%%%%%%%%%%%%%%%%%%%%%%%%%%%%%%%%%%%%%%%%%%%%%%%%%%%%%
\section{Proposal}

%%%%%%%%%%%%%%%%%%%%%%%%%%%%%%%%%%%%%%%%
\subsection{Adaptive T-soft update}

The first two of the three issues mentioned above are resolved by deriving a new update rule, so-called AT-soft update.
To develop AT-soft update, the formulation of AdaTerm~\cite{ilboudo2022adaterm}, which is a kind of stochastic gradient descent method with the adaptive noise robustness, is applied.
In this method, by assuming that the gradient is generated from student-t distribution, its location, scale, and degrees of freedom parameters, which are utilized for updating the network, can be estimated by approximate maximum likelihood estimation.
Instead of the gradient as stochastic variable, the parameters of the main network are considered to be generated from student-t distribution, and its location is mapped to the parameters of the target network.
With this assumption, AT-soft update obtains the noise robustness as in the conventional T-soft update.
In addition, the degrees of freedom can be estimated at the same time in this formulation, so that the noise robustness can be automatically adjusted according to the faced task.

Specifically, the $i$-th subset of $\theta_{V, \pi}$ (e.g. a weight matrix in each layer), $\theta_i$ with $d_i$ the number of dimensions, is assumed to be generated from $d_i$-dimensional diagonal student-t distribution with three kinds of sample statistics:
a location parameter $\bar{\theta}_i \in \mathbb{R}^{d_i}$;
a scale parameter $\sigma_i \in \mathbb{R}_{+}^{d_i}$;
and degrees of freedom $\nu_i \in \mathbb{R}_{+}$.
With $\tilde{\nu}_i = \nu_i d_i^{-1}$ and $D_i = d_i^{-1} \sum_{j=1}^{d_i} (\theta_{i,j} - \bar{\theta}_{i,j})^2 \sigma_{i,j}^{-2}$, its density can be described as below.
\begin{align}
    \theta_i &\sim \cfrac{\Gamma(\frac{\nu_i + d_i}{2})}{\Gamma(\frac{\nu_i}{2})(\nu_i \pi)^{\frac{d_i}{2}} \prod_{j=1}^{d_i} \sigma_{i,j}}
    \left (1 + \cfrac{D_i}{\tilde{\nu}_i} \right )^{-\frac{\nu_i + d}{2}}
    \nonumber \\
    &=: \mathcal{T}(\theta_i \mid \bar{\theta}_i, \sigma_i, \nu_i)
\end{align}
where $\Gamma$ denotes the gamma function.
Note that the conventional T-soft update simplifies this model as one-dimensional student-t distribution for the mean of $\theta_i - \bar{\theta}_i$, but here we treat it as $d_i$-dimensional distribution with slightly increased computational cost.

With this assumption, following the derivation of AdaTerm, $\bar{\theta}_i$, $\sigma_i$, and $\nu_i$ are optimally inferred to maximize the approximated log-likelihood.
The important variable in the derivation is $w_1$, which indicates the deviation of $\theta_i$ from $\mathcal{T}$, and is calculated as follows:
\begin{align}
    w_1 = \cfrac{\tilde{\nu}_i + 1}{\tilde{\nu}_i + D_i}
\end{align}
That is, since $D_i$ represents the pseudo-distance from $\mathcal{T}$, the larger $D_i$ is, the closer $w_1$ is to $0$.
In addition, the smaller $\tilde{\nu}_i$ is, the more sensitive $w_1$ is to fluctuations in $D_i$, leading to higher noise robustness.
Using $w_1$, $w_2$, which is used only for updating $\tilde{\nu}_i$, can be derived as follows:
\begin{align}
    w_2 = w_1 - \ln(w_1)
\end{align}

These $w_{1,2}$ are used to calculate the update ratio of the sample statistics, $\tau_{1,2}$.
\begin{align}
    \tau_{1,2} = \tau \cfrac{w_{1,2}}{\overline{w}_{1,2}}
\end{align}
where $\tau \in (0, 1]$ denotes the basic update ratio given as a hyperparameter.
To satisfy $\tau_{1,2} \in (0, 1]$, the upper bounds of $w_{1,2}$, $\overline{w}_{1,2}$, are employed.
\begin{align}
    \overline{w}_1 = \cfrac{\tilde{\nu}_i + 1}{\tilde{\nu}_i}
    , \
    \overline{w}_2 = \max(\overline{w}_1 - \ln(\overline{w}_1), 87.3365)
\end{align}
where $87.3365$ means the negative logarithm with the tiny number of float32.

The update amounts for $\bar{\theta}_i$, $\sigma_i^2$, and $\tilde{\nu}_i$ are respectively given as follows:
\begin{align}
    \bar{\theta}_i^\prime &= \theta_i
    \\
    (\sigma_i^2)^\prime &= (\theta_i - \bar{\theta}_i)^2
    \nonumber \\
    &+ \max(\epsilon^2, ((\theta_i - \bar{\theta}_i)^2 - D_i \sigma_i^2)\tilde{\nu}_i^{-1})
    \\
    \tilde{\nu}_i^\prime &= \left( \cfrac{\tilde{\nu}_i + 2}{\tilde{\nu}_i + 1} + \tilde{\nu}_i \right) \cfrac{\tilde{\nu}_i - \underline{\tilde{\nu}}}{\tilde{\nu}_i w_2} + \underline{\tilde{\nu}} + \epsilon
\end{align}
where $\epsilon \ll 1$ denotes the small value for stabilizing the computation and $\underline{\tilde{\nu}}$ denotes the lower bound of $\tilde{\nu}$ (i.e. the maximum noise robustness) given as a hyperparamter.

Using the update ratios and the update amounts obtained above, $\bar{\theta}_i$, $\sigma_i^2$, and $\tilde{\nu}_i$ can be updated.
\begin{align}
    \bar{\theta}_i &\gets (1 - \tau_1) \bar{\theta}_i + \tau_1 \bar{\theta}_i^\prime
    \\
    \sigma_i^2 &\gets (1 - \tau_1) \sigma_i^2 + \tau_1 (\sigma_i^2)^\prime
    \\
    \tilde{\nu}_i &\gets (1 - \tau_2) \tilde{\nu}_i + \tau_2 \tilde{\nu}_i^\prime
\end{align}
As a result, AT-soft update enables to update the parameters set $\bar{\theta}_i$ of the target network adaptively (i.e. depending on the deviation of $\theta_i$ from $\mathcal{T}$), while automatically tuning the noise robustness represented by $\sigma_i^2$ and $\tilde{\nu}_i$.

%%%%%%%%%%%%%%%%%%%%%%%%%%%%%%%%%%%%%%%%
\subsection{Consolidation from main to target networks}

However, if $\tau_1 \simeq 0$ continues in the above update, $\theta_i$ will deviate from $\bar{\theta}_i$ gradually, and the target network will no longer be able to generate pseudo-supervised signals since the assumption $\bar{\theta}_i \simeq \theta_i$ is broken.
In such a case, parts of $\theta_i$ would be updated with the minimization of eqs.~\eqref{eq:loss_value}--\eqref{eq:loss_policy} in a wrong direction, causing $\tau_1 \simeq 0$ as outlier.
Hence, to stop this fruitless judgement and restart the appropriate updates, reverting $\theta_i$ to $\bar{\theta}_i$ and holding $\theta_i \simeq \bar{\theta}_i$ would be the natural effective way.
To this end, a heuristic consolidation is designed as below.

Specifically, the update ratio from the main to target networks, $\tau_c$, is designed to be larger when the update ratio of $\bar{\theta}_i$ is smaller (i.e. when $\theta_i$ deviates from $\mathcal{T}$).
\begin{align}
    \tau_c = \lambda \tau \left ( 1 - \cfrac{w_1}{\overline{w}_1} \right )
\end{align}
where $\lambda \in [0, 1]$ adjusts the strength of this consolidation, which should be the same as or weaker than the update speed of the target network.

Next, since consolidating all $\theta_i$ would interfere with learning, the consolidated subset of outliers, $\theta_i^c$, should be extracted.
The simple and popular way is to use the $q$-th quantile $Q$ with $q \in [0, 1]$.
Since the component that contributes to making $w_1$ small is with large $(\theta_{i,j} - \bar{\theta}_{i,j})^2 \sigma_{i,j}^{-2} =: \Delta_{i,j}$, $\theta_i^c$ is defined as follows:
\begin{align}
    \theta_i^c = \left \{ \theta_{i,j} \in \theta_i \mid \Delta_{i,j} \geq Q(\Delta_i; q) \right \}
\end{align}

\begin{figure}[tb]
    \centering
    \includegraphics[keepaspectratio=true,width=0.96\linewidth]{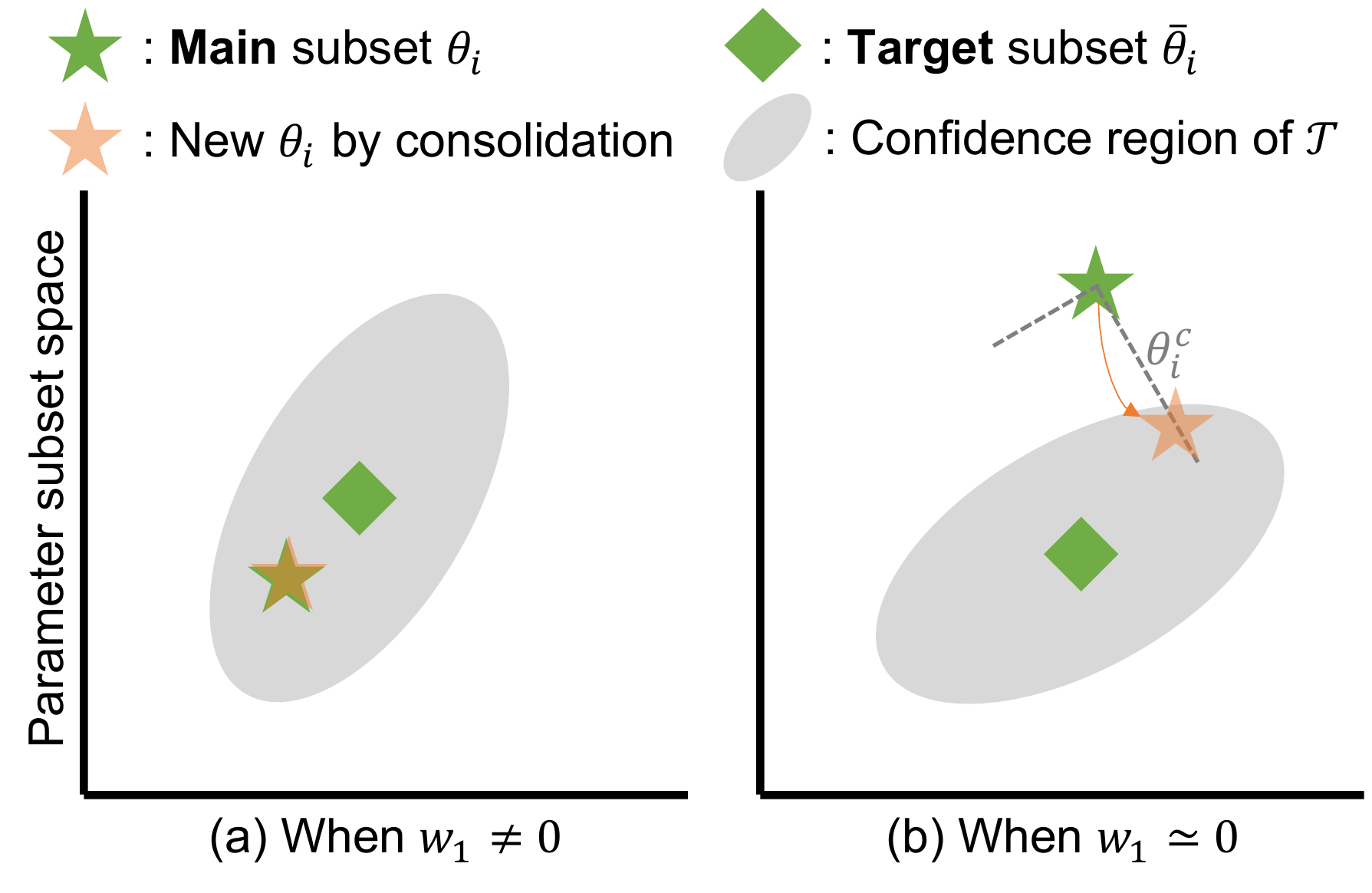}
    \caption{Rough sketches of the proposed consolidation:
    when $\theta_i$ is not far from $\mathcal{T}$ as in (a), almost no consolidation works;
    when parts of $\theta_i$ are far from $\mathcal{T}$ as in (b), only $\theta_i^c$, which leads to $w_1 \simeq 0$, is consolidated to the target network.
    }
    \label{fig:img_consolidate}
\end{figure}

Thus, the following update formula consolidates $\theta_i^c$ to the corresponding subset of the target network, $\bar{\theta}_i^c$.
\begin{align}
    \theta_i^c \gets (1 - \tau_c) \theta_i^c + \tau_c \bar{\theta}_i^c
\end{align}
A rough sketch of this consolidation is shown in Fig.~\ref{fig:img_consolidate}.
Although loss-function-based consolidations, as proposed in the context of continual learning~\cite{kirkpatrick2017overcoming,zenke2017continual}, would be also possible, a more convenient implementation with lower computational cost was employed.

The pseudo-code for the consolidated adaptive T-soft (CAT-soft) update is summarized in Alg.~\ref{alg:catsoft}.
Note that, although $\underline{\tilde{\nu}}$ must be specified as a new hyperparameter in (C)AT-soft update, $\nu$ specified in T-soft update is already conservatively set for noise, and we can inherit it (i.e. $\underline{\tilde{\nu}} = 1$).
Therefore, the additional hyperparameters to be tuned are $\lambda$ and $q$.
$q$ can be given as $q \simeq 1$ so that a few parameters in $i$-th subset are consolidated without interfering with learning.
$\lambda$ can be given as the inverse of the number of parameters to be consolidated.
In other words, we can decide whether to make $q$ closer to $1$ and consolidate fewer parameters tightly, or make $q$ smaller and consolidate more parameters slightly.

%Figure

%Algorithm
\begin{algorithm}[tb]
    \caption{CAT-soft update}
    \label{alg:catsoft}
    \begin{algorithmic}[1]
        \State{(Initialize $\bar{\theta}_i = \theta_i$, $\sigma_i = \epsilon$, $\tilde{\nu}_i = \underline{\tilde{\nu}}$)}
        \For{$\theta_i, \bar{\theta}_i \subset \theta_{V, \pi}, \bar{\theta}_{V, \pi}$}
            \State{$\Delta_i = (\theta_i - \bar{\theta}_i)^2 \sigma_i^{-2}$}
            \State{$D_i = d_i^{-1} \sum_{j=1}^{d_i} \Delta_{i,j}$}
            \State{$w_1 = (\tilde{\nu}_i + 1)(\tilde{\nu}_i + D_i)^{-1}$,
                $w_2 = w_1 - \ln(w_1)$}
            \State{$\overline{w}_1 = (\tilde{\nu}_i + 1)\tilde{\nu}_i^{-1},
                \overline{w}_2 = \max(\overline{w}_1 - \ln(\overline{w}_1), 87.3365)$}
            \State{$\tau_{1,2} = \tau w_{1,2}\overline{w}_{1,2}^{-1}$}
            \State{$(\sigma_i^2)^\prime = \Delta_i \sigma_i^2
                + \max(\epsilon^2, (\Delta_i - D_i) \sigma_i^2 \tilde{\nu}_i^{-1})$}
            \State{$\tilde{\nu}_i^\prime = \{1 + (\tilde{\nu}_i + 1)^{-1} + \tilde{\nu}_i\}
                (\tilde{\nu}_i - \underline{\tilde{\nu}})(\tilde{\nu}_i w_2)^{-1}
                + \underline{\tilde{\nu}} + \epsilon$}
            \State{$\bar{\theta}_i \gets (1 - \tau_1) \bar{\theta}_i + \tau_1 \theta_i$}
            \State{$\sigma_i^2 \gets (1 - \tau_1) \sigma_i^2 + \tau_1 (\sigma_i^2)^\prime$}
            \State{$\tilde{\nu}_i \gets (1 - \tau_2) \tilde{\nu}_i + \tau_2 \tilde{\nu}_i^\prime$}
            \If{Consolidation}
                \State{$\tau_c = \lambda \tau ( 1 - w_1\overline{w}_1^{-1})$}
                \State{$\theta_i^c = \{ \theta_{i,j} \in \theta_i \mid \Delta_{i,j} \geq Q(\Delta_i; q) \}$}
                \State{$\theta_i^c \gets (1 - \tau_c) \theta_i^c + \tau_c \bar{\theta}_i^c$}
            \EndIf
        \EndFor
    \end{algorithmic}
\end{algorithm}

%%%%%%%%%%%%%%%%%%%%%%%%%%%%%%%%%%%%%%%%%%%%%%%%%%%%%%%%%%%%%%%%%%%%%%%%%%%%%%%%
\section{Simulations}

%Table
\begin{table}[tb]
    \caption{Hyperparameters for the used DRL algorithms}
    \label{tab:param}
    \centering
    \begin{tabular}{ccc}
        \hline\hline
        Symbol & Meaning & Value
        \\
        \hline
        $L$ & \#Hidden layer & $2$
        \\
        $N$ & \#Neuron for each layer & $100$
        \\
        $\gamma$ & Discount factor & $0.99$
        \\
        $(\alpha, \beta, \epsilon, \underline{\tilde{\nu}})$ & For AdaTerm~\cite{ilboudo2022adaterm} & $(10^{-3}, 0.9, 10^{-5}, 1)$
        \\
        $(\kappa, \beta, \lambda, \underline{\Delta})$ & For PPO-RPE~\cite{kobayashi2021proximal} & $(0.5, 0.5, 0.999, 0.1)$
        \\
        $(N_c, N_b, \alpha, \beta)$ & For PER~\cite{schaul2015prioritized} & $(10^4, 32, 1.0, 0.5)$
        \\
        $(\sigma, \underline{\lambda}, \overline{\lambda}, \beta)$ & For L2C2~\cite{kobayashi2022l2c2} & $(1, 0.01, 1, 0.1)$
        \\
        \hline\hline
    \end{tabular}
\end{table}

%%%%%%%%%%%%%%%%%%%%%%%%%%%%%%%%%%%%%%%%
\subsection{Setup}

For the statistical verification of the proposed method, the following simulations are conducted.
As simulation environments, Pybullet~\cite{coumans2016pybullet} with OpenAI Gym~\cite{brockman2016openai} is employed.
From it, \textit{InvertedDoublePendulumBulletEnv-v0} (DoublePendulum), \textit{HopperBulletEnv-v0} (Hopper), and \textit{AntBulletEnv-v0} (Ant) are chosen as tasks.
To make the tasks harder, the observations from them are with white noises, scale of which is $0.001$.
With 18 random seeds, each task is tried to be accomplished by each method.
After training, the learned policy is run 100 times for evaluating the sum of rewards as a score (larger is better).

The implementation of the base network architecture and DRL algorithm is basically the same as in the literature~\cite{kobayashi2022l2c2}.
However, it is noticeable that the stochastic policy function is modeled by student-t distribution for conservative learning and efficient exploration~\cite{kobayashi2019student}, instead of normal distribution.
Hyperparamters for the implementation are summarized in Table~\ref{tab:param}.
Note that the learning difficulty is higher because the specified value of the learning rate is higher than one suitable for DRL, revealing the effectiveness of the target network for stabilizing learning.

The following three methods are compared.
\begin{itemize}
    \item T-soft update: $(\tau=0.1, \nu=1)$
    \item AT-soft update: $(\tau=0.1, \underline{\tilde{\nu}}=1)$
    \item CAT-soft update: $(\tau=0.1, \underline{\tilde{\nu}}=1, \lambda=1, q=1)$
\end{itemize}
Here, $q$ is designed to consolidate only one parameter in each subset as the simplest implementation.
Correspondingly, $\lambda$ is given as the inverse of the (maximum) number of consolidated parameters.
$\tau=0.1$ is set smaller than one in the literature~\cite{kobayashi2021t}, but this is to counteract the negative effects of the high learning rate (i.e. $10^{-3}$) set above.

%Figure
\begin{figure*}[tb]
    \begin{subfigure}[b]{0.32\linewidth}
        \centering
        \includegraphics[keepaspectratio=true,width=\linewidth]{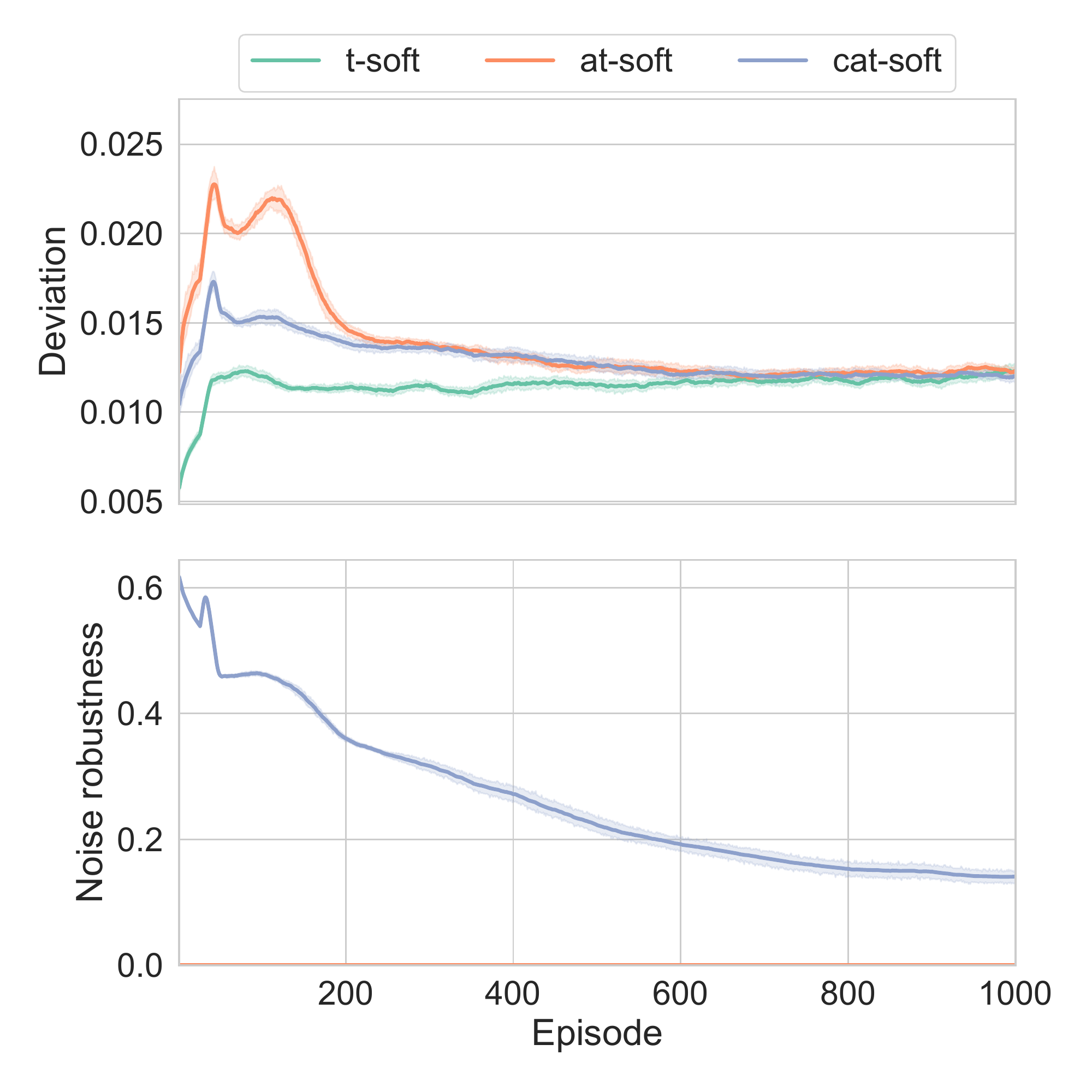}
        \subcaption{DoublePendulum}
        \label{fig:DoublePendulum}
    \end{subfigure}
    \begin{subfigure}[b]{0.32\linewidth}
        \centering
        \includegraphics[keepaspectratio=true,width=\linewidth]{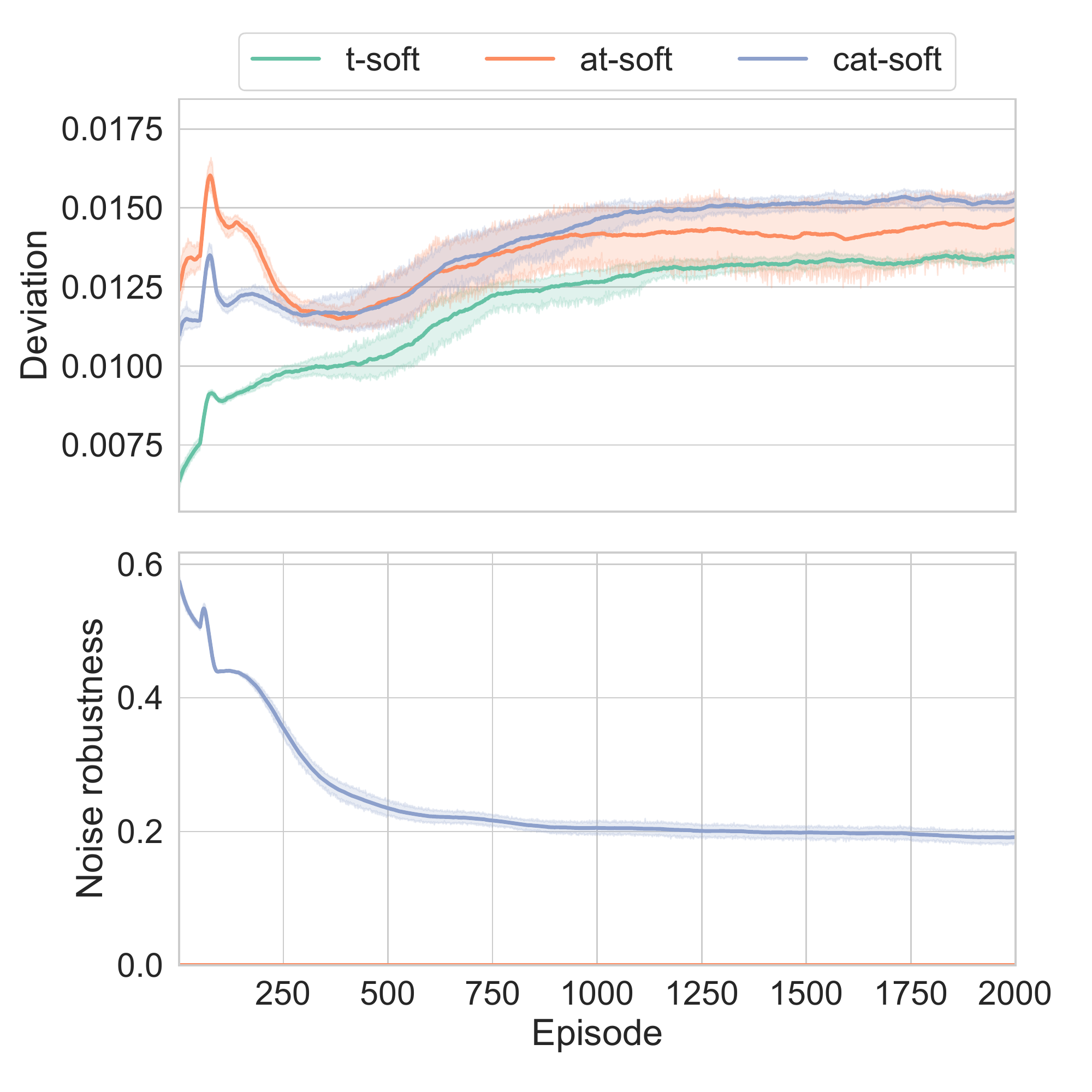}
        \subcaption{Hopper}
        \label{fig:Hopper}
    \end{subfigure}
    \begin{subfigure}[b]{0.32\linewidth}
        \centering
        \includegraphics[keepaspectratio=true,width=\linewidth]{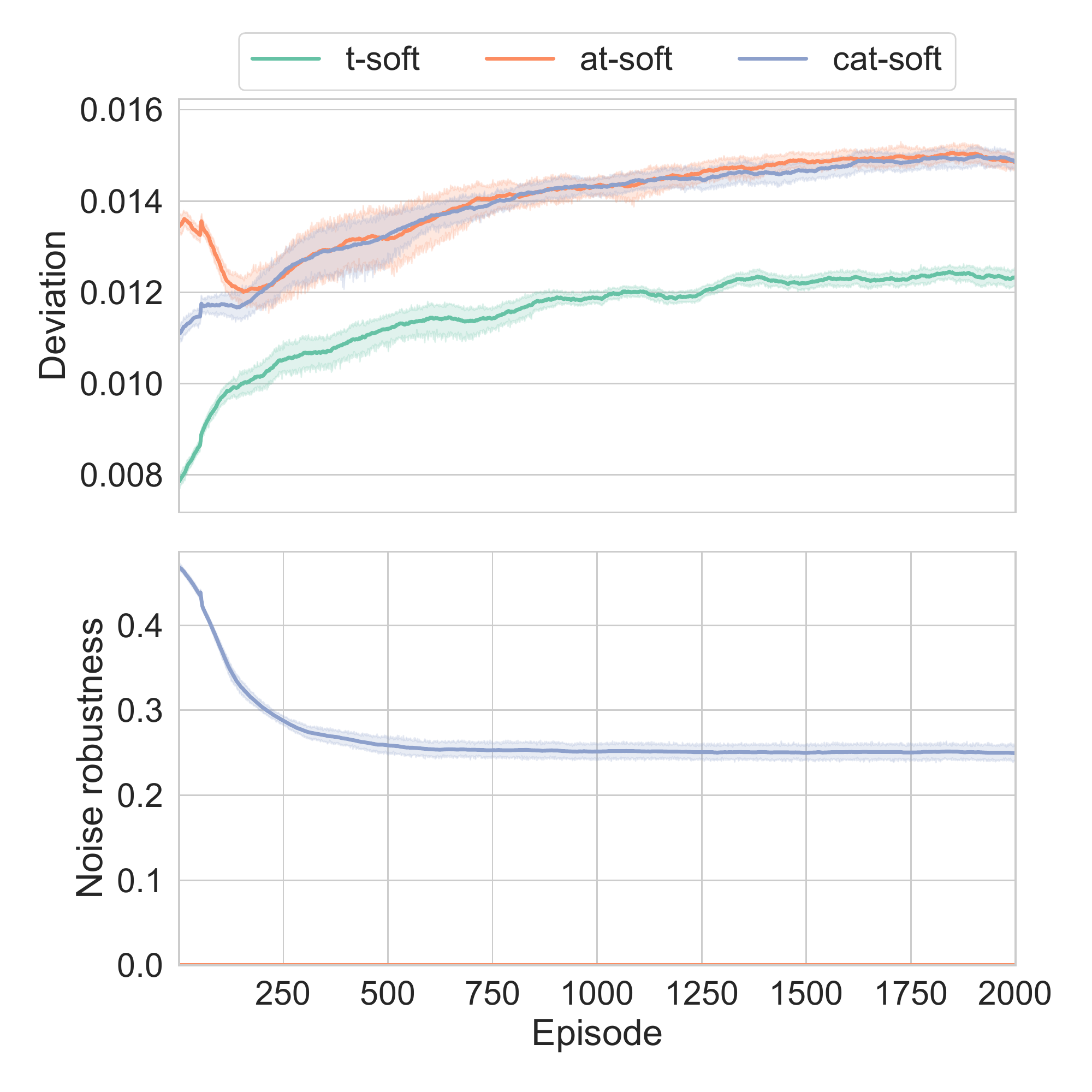}
        \subcaption{Ant}
        \label{fig:Ant}
    \end{subfigure}
    \caption{Learning curves for benchmarks:
    the upper row shows mean of the deviation between the main and target networks, $|\theta - \bar{\theta}|$;
    the lower row plots mean of $1 - w_1 / \bar{w}_1$, which corresponds to the noise robustness (or the magnitude of consolidation);
    compared to T-soft update, AT-soft and CAT-soft updates have the larger deviation, but this was suppressed by the consolidation in the early stages of learning;
    as learning progresses, the cases, where updates of the target network were suppressed due to noise, were decreased, and the consolidation was relaxed, resulting in AT-soft and CAT-soft updates converging to roughly the same degree of deviation.
    }
    \label{fig:sim_learn}
\end{figure*}

%Table
\begin{table}[tb]
    \caption{The sum of rewards after training}
    \label{tab:sim_score}
    \centering
    \begin{tabular}{l ccc}
        \hline\hline
        Method & DoublePendulum & Hopper & Ant
        \\
        \hline
        T-soft
        & 6427.1 $\pm$ 3357.8
        & 1852.8 $\pm$ 900.9
        & 2683.8 $\pm$ 249.3
        \\
        AT-soft
        & 6379.7 $\pm$ 3299.7
        & 1662.7 $\pm$ 897.4
        & 2764.1 $\pm$ 265.5
        \\
        CAT-soft
        & 7129.2 $\pm$ 2946.0
        & 1971.2 $\pm$ 812.9
        & 2760.0 $\pm$ 312.2
        \\
        \hline\hline
    \end{tabular}
\end{table}

%%%%%%%%%%%%%%%%%%%%%%%%%%%%%%%%%%%%%%%%
\subsection{Result}

The learning behaviors are depicted in Fig.~\ref{fig:sim_learn}.
As pointed out, the deviation by (C)AT-soft updates were larger than that of the conventional T-soft update since (C)AT-soft update have better outlier and noise detection performance and the target network update is easily suppressed.
This was pronounced in the early stage of training when the noise robustness is high and the update of the main network is unstable.
However, CAT-soft update suppressed the deviation in the early stage of training.
As the learning progresses, CAT-soft update converged to roughly the same level of deviation as AT-soft update because the consolidation was relaxed with the weakened noise robustness.

The scores of 100 runs after learning are summarized in Table~\ref{tab:sim_score}.
AT-soft update slightly increased the performance of T-soft update on Ant, but decreased it on Hopper.
In contrast, CAT-soft update outperformed T-soft update in all tasks.

%%%%%%%%%%%%%%%%%%%%%%%%%%%%%%%%%%%%%%%%
\subsection{Demonstration}

%Table
\begin{table}[tb]
    \caption{Arguments for \textit{MinitaurBulletDuckEnv-v0}}
    \label{tab:minitaur}
    \centering
    \begin{tabular}{ccc}
        \hline\hline
        Argument & Default & Modified
        \\ \hline
        \texttt{motor\_velocity\_limit} & $\infty$ & 100
        \\
        \texttt{pd\_control\_enabled} & False & True
        \\
        \texttt{accurate\_motor\_model\_enabled} & True & False
        \\
        \texttt{action\_repeat} & 1 & 4
        \\
        \texttt{observation\_noise\_stdev} & 0 & $10^{-5}$
        \\
        \texttt{hard\_reset} & True & False
        \\
        \texttt{env\_randomizer} & Uniform & None
        \\
        \texttt{distance\_weight} & 1 & 100
        \\
        \texttt{energy\_weight} & 0.005 & 1
        \\
        \texttt{shake\_weight} & 0 & 1
        \\
        \hline\hline
    \end{tabular}
\end{table}

%Figure
\begin{figure}[tb]
    \begin{subfigure}[b]{0.96\linewidth}
        \centering
        \includegraphics[keepaspectratio=true,width=\linewidth]{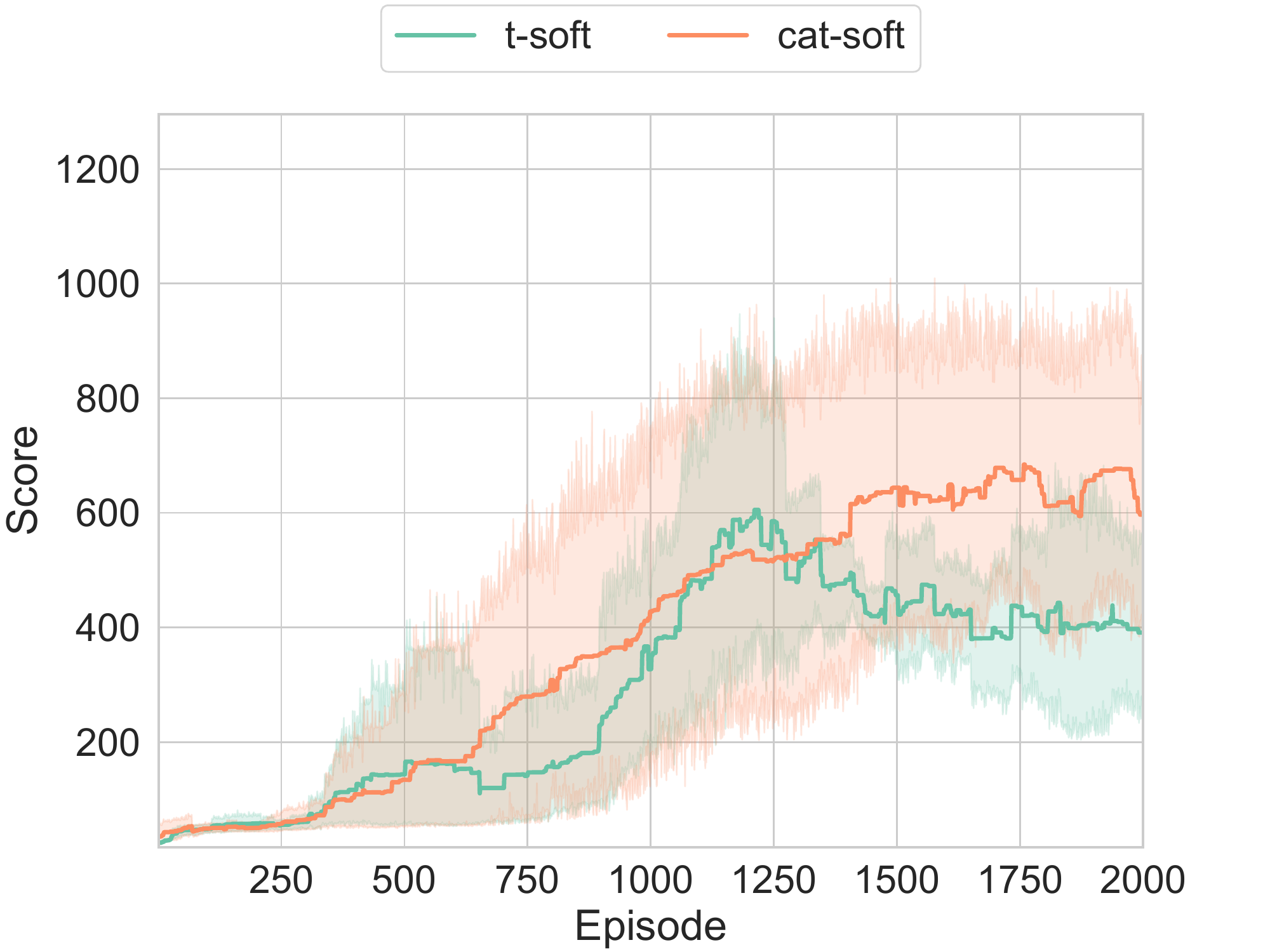}
        \subcaption{Learning curves}
        \label{fig:demo_learn}
    \end{subfigure}
    \begin{subfigure}[b]{0.96\linewidth}
        \centering
        \includegraphics[keepaspectratio=true,width=\linewidth]{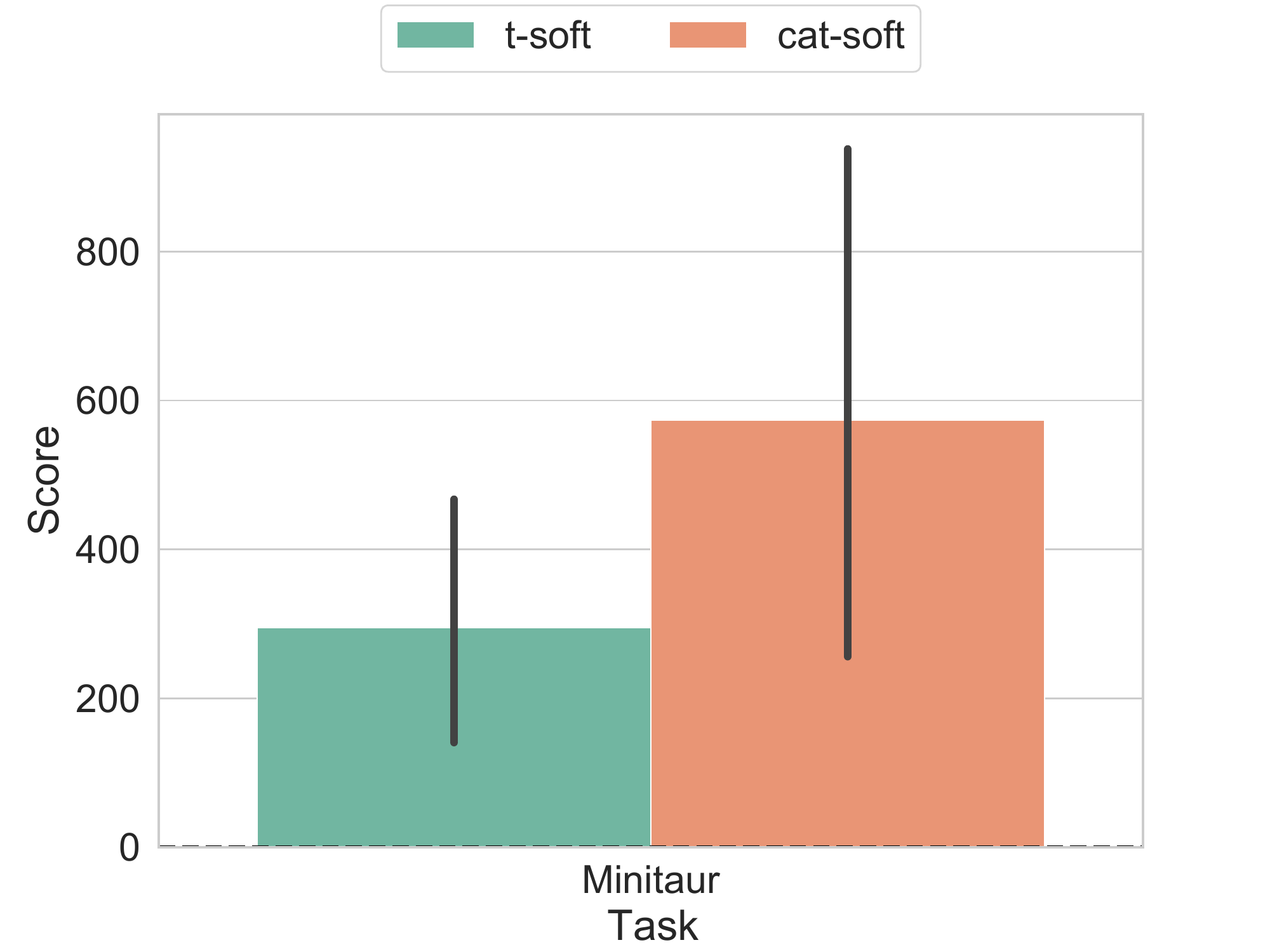}
        \subcaption{Test results}
        \label{fig:demo_score}
    \end{subfigure}
    \caption{Demonstration results:
    CAT-soft update showed the more stable learning curve than that of T-soft update and reached the success level of the task in the final performance.
    }
    \label{fig:demo}
\end{figure}

As a demonstration, a simulation closer to the real robot experiment, \textit{MinitaurBulletDuckEnv-v0} (Minitaur) in Pybullet, is tried.
The task is to move a duck on top of a Ghost Minitaur, a quadruped robot developed by Ghost Robotics.
Since this duck is not fixed, careful locomotion is required, and its states (e.g. position) are unobserved, making this task a partially observed MDP (POMDP).
Note that the default setting for Minitaur tasks is unrealistic, as pointed out in the literature~\cite{kobayashi2021optimistic}.
Therefore, it was modified as shown in Table~\ref{tab:minitaur} (arguments not listed are left at default).

T-soft and CAT-soft updates are compared under the same conditions as in the above simulations.
The learning curves of the scores for 8 trials and the test results of the trained policies are depicted in Fig.~\ref{fig:demo}.
The best behaviors on the tests can be found in the attached video.
As can be seen from Fig.~\ref{fig:demo}, only the proposed CAT-soft update was able to acquire the successful cases of the task (walking without dropping the duck).
Thus, it is suggested that CAT-soft update can contribute to the success of the task by steadily improving the learning performance even for more practical tasks.

%%%%%%%%%%%%%%%%%%%%%%%%%%%%%%%%%%%%%%%%%%%%%%%%%%%%%%%%%%%%%%%%%%%%%%%%%%%%%%%%
\section{Conclusion}

This paper proposed a new update rule for the target network, CAT-soft update, which stabilizes DRL.
In order to adaptively adjust the noise robustness, the update rule inspired by AdaTerm, which has been developed recently, was derived.
In addition, a heuristic consolidation from the main to target networks was developed to suppress the deviation between them, which may occur when updates are continuously limited due to noise.
The developed CAT-soft update was tested on the DRL benchmark tasks, and succeeded in improving and stabilizing the learning performance over the conventional T-soft update.

Actually, the target network should not deviate from the main network in terms of its outputs, not in terms of its parameters.
A new consolidation and a noise-robust update based on the output space are expected to contribute to further performance improvements.
These efforts to stabilize  DRL will lead to its practical use in the near future.

%%%%%%%%%%%%%%%%%%%%%%%%%%%%%%%%%%%%%%%%%%%%%%%%%%%%%%%%%%%%%%%%%%%%%%%%%%%%%%%%
% \section*{APPENDIX}
%
%
%
%%%%%%%%%%%%%%%%%%%%%%%%%%%%%%%%%%%%%%%%%%%%%%%%%%%%%%%%%%%%%%%%%%%%%%%%%%%%%%%%
\section*{ACKNOWLEDGMENT}

This work was supported by JSPS KAKENHI, Grant-in-Aid for Scientific Research (B), Grant Number JP20H04265.

%%%%%%%%%%%%%%%%%%%%%%%%%%%%%%%%%%%%%%%%%%%%%%%%%%%%%%%%%%%%%%%%%%%%%%%%%%%%%%%%
\bibliographystyle{IEEEtran}
{
\bibliography{biblio}
}

\end{document}